\documentclass[letterpaper, 10 pt, conference]{ieeeconf}  

\IEEEoverridecommandlockouts                              

\usepackage{hyperref}       
\usepackage{url}            
\usepackage{booktabs}       
\usepackage{amsfonts}       
\usepackage{nicefrac}       
\usepackage{microtype}      
\usepackage{xcolor}         

\usepackage{graphicx}
\usepackage[misc]{ifsym}
\usepackage{amsmath}
\usepackage{amssymb}
\usepackage{multirow}
\usepackage{makecell}
\usepackage{threeparttable}
\usepackage{lipsum}
\usepackage{tabularx}
\usepackage{diagbox}
\usepackage{color}
\usepackage{enumerate}
\usepackage{wrapfig}
\usepackage{multicol}
\usepackage{subfigure}
\usepackage{bbding}

\begin{document}

\title{\LARGE \bf Boosting 3D Point Cloud Registration by Transferring Multi-modality Knowledge}

\author{Mingzhi Yuan$^{1\dagger}$, Xiaoshui Huang$^{2\dagger}$, Kexue Fu$^{1\dagger}$, Zhihao Li$^{1}$ and Manning Wang$^{1}\textsuperscript{\Envelope}$
\thanks{$^{1}$Mingzhi Yuan, Kexue Fu, Zhihao Li and Manning Wang are with the Digital Medical Research Center, School of Basic Medical Science, Fudan University, Shanghai 200032, China, and also with the Shanghai Key Laboratory of Medical Image Computing and Computer Assisted Intervention, Shanghai, China, 200032, {\tt\small \{mzyuan20, fukexue,  mnwang\}@fudan.edu.cn}, {\tt\small lizhihao21@m.fudan.edu.cn}}%
\thanks{$^{2}$Xiaoshui Huang is with the Shanghai artificial intelligence Lab. {\tt\small \{huangxiaoshui\}@pjlab.org.cn}}%
\thanks{$^{\dagger}$ These authors contributed equally. }%
\thanks{$\textsuperscript{\Envelope}$ Corresponding author.}%
}

\maketitle
\thispagestyle{empty}
\pagestyle{empty}

\begin{abstract}
The recent multi-modality models have achieved great performance in many vision tasks because the extracted features contain the multi-modality knowledge. However, most of the current registration descriptors have only concentrated on local geometric structures. This paper proposes a method to boost point cloud registration accuracy by transferring the multi-modality knowledge of pre-trained multi-modality model to a new descriptor neural network. Different to the previous multi-modality methods that requires both modalities, the proposed method only requires point clouds during inference. Specifically, we propose an ensemble descriptor neural network combining pre-trained sparse convolution branch and a new point-based convolution branch. By fine-tuning on a single modality data, the proposed method achieves new state-of-the-art results on 3DMatch and competitive accuracy on 3DLoMatch and KITTI. The code and the trained model will be released at \url{https://github.com/phdymz/DBENet.git}.
\end{abstract}

\section{Introduction}
Multi-modality data has been demonstrated inspiring performance in numerous vision tasks. Typical examples are the recent unsupervised general  models, such as CLIP \cite{radford2021learning}, Flangmigo \cite{alayrac2022flamingo}, which achieve accurate and high generalization performance in vision tasks. The multi-modality model usually contains plentiful of knowledge from several vision modalities. However, training such a general model requires tremendous computational and storage resources. It is usually difficult for common researchers to train such a model. Developing a multi-modality model  becomes much more challenging in the 3D computer vision as the 3D multi-modality data acquisition is very expensive. In this paper, we focus on a specific task by investigating how to use the multi-modality information to solve the point cloud registration. We are not going to train such a multi-modality model. We propose a method to transfer the knowledge of an existing multi-modality model to the point cloud registration task.
 
\begin{figure}[ht]
  \centering
  \includegraphics[width=0.48\textwidth]{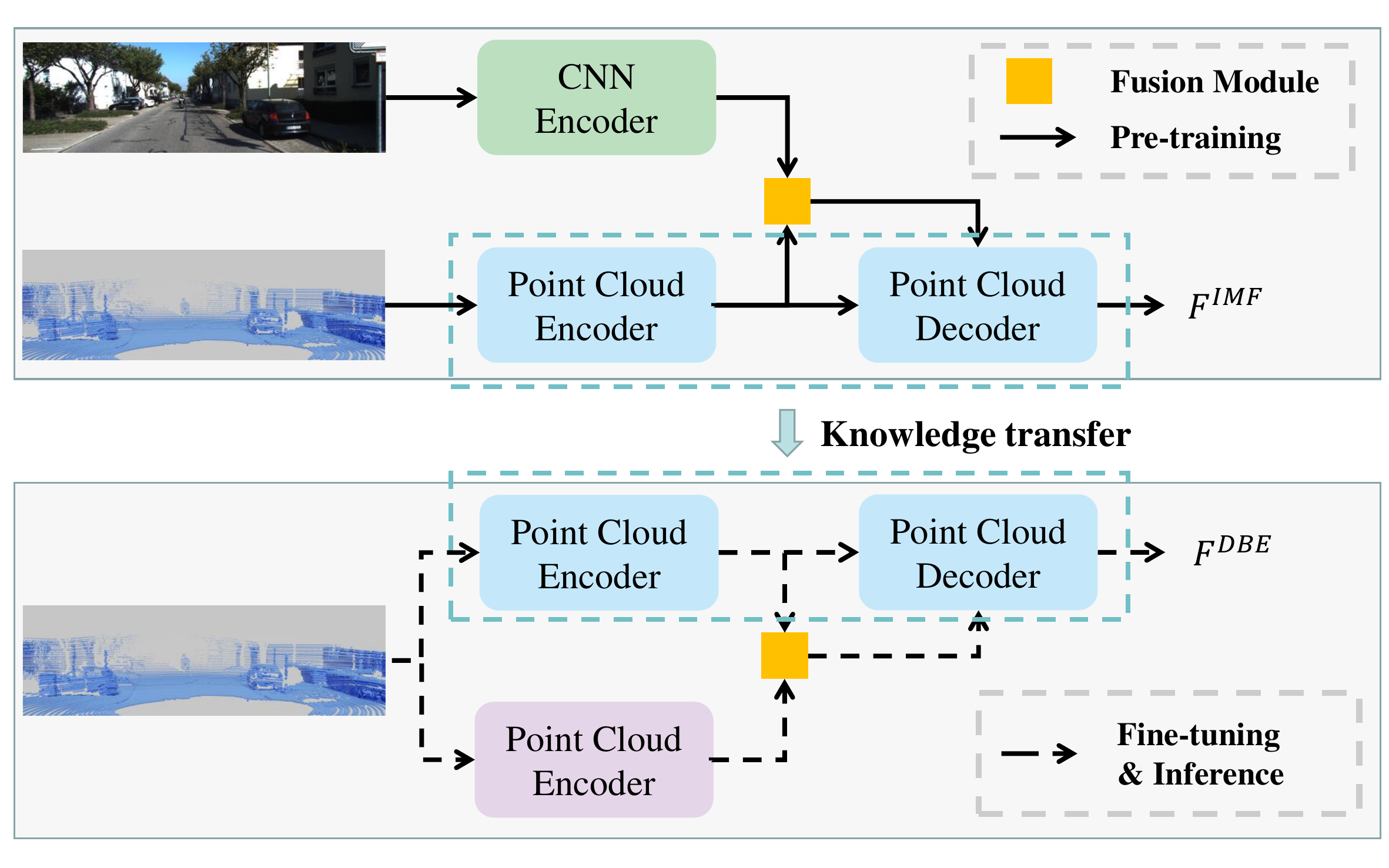}
  \caption{
  \textbf{The overall architecture of our proposed method.} 
  The network above is a dual-branch network pre-trained on multiple modalities. 
  The network below is a dual-branch network fine-tuned on point cloud. 
  We transfer the knowledge learned from multiple modalities to the network below by transferring the weights of pre-trained point cloud branch. 
  Benefiting from knowledge transfer, the network below can achieve comparable performance to the above network but with only single modality input at inference. $F^{IMF}$ and $F^{DBE}$ both represent the feature matrix for point cloud registration. 
  }
  \label{fig1}
  \vspace{-0.3em}
\end{figure}

3D point cloud registration \cite{ref1} aims at estimating a transformation between two unaligned point clouds, which is critical to many applications including robotics \cite{ref2}, autonomous driving \cite{ref3}, and SLAM \cite{ref4}. 
Current state-of-the-art methods \cite{ref5,ref6} commonly start from deep feature extraction and matching, followed by robust model fitting methods e.g. RANSAC \cite{ref7} for robust transformation estimation. Although a series of distinctive deep features \cite{ref8,ref29,ref33,ref34} have been proposed recently, point cloud registration in practical scenes remains challenging due to reliance on geometric information and ignorance of semantic information.

In the current research stream of point cloud registration, plenty of deep descriptors \cite{ref5,ref6,ref8,ref9} only concentrate on local geometric structures. However, in practical applications, there may exist many repeatable and ambiguous structures in point clouds, such as floors, ceilings, and walls, which tend to lead to wrong matches. A feasible solution is to incorporate extra semantic information to make descriptors more reliable. 
Recently, IMFNet \cite{ref10} introduced an additional image encoder and fused geometric information in point cloud with semantic information in image using a cross-attention. 
Since features generated from two different modalities have complementary information, IMFNet successfully improved the original pure point cloud descriptor by a large margin. 
However, it’s not easy to take multi-modality data as input since IMFNet requires both RGB and point cloud during inference. 
Specifically, it’s tedious to deal with spatial and temporal calibration and synchronization for different sensors. 
Moreover, acquiring multi-modality data will inevitably reduce the fault tolerance of the system, since the breakdown of either sensor can lead to failure of registration. 

To tackle above problem, an intuitive solution is to transfer the knowledge learned from multiple modalities to single modality. 
We find that a point cloud branch in multi-modality network trained on multi-modality data extracts richer information than a single point cloud network trained on pure point cloud data. 
Therefore, in this paper, we propose an approach for 3D point cloud registration, which transfers the knowledge of existing pre-trained multi-modality model to a new neural network that only needs point clouds during inference. 
As shown in Figure \ref{fig1}, our method utilizes two dual-branch networks. 
The network above is a existing pre-trained network, which takes both point cloud and image as input. 
The network below retains the pre-trained point cloud branch, and replace the image branch with an extra point cloud network. 
The reserved branch contains the knowledge learned from multiple modalities and the other modules are fine-tuned using single modality data (point cloud). 
During inference, the fine-tuned network only takes point clouds as input and achieves better performance by considering the complementary information in two point cloud branches.

The main contributions of our work are:
\begin{itemize}
    \item We propose a framework transferring knowledge learned from multi-modality pre-training, which improves the performance of 3D point cloud registration while avoiding complex calibration and synchronization of sensors at inference time.
    \item We provide analysis of the different strategies for knowledge transfer between point cloud and image and conduct experiments to verify it.
    \item We propose a learnable dual-branches ensemble descriptor for point cloud registration, which consists of two mainstream feature extractors. The ablation shows that it’s more powerful than a single feature extractor. 
    \item Our method achieves new state-of-the-art results on 3DMatch \cite{ref11} and shows competitive performances on 3DLoMatch \cite{ref5} and Kitti \cite{ref12}.
\end{itemize}

\section{Related works}
\subsection{Point cloud registration}
The methods for point cloud registration can be roughly divided into two categories: correspondence-free methods and correspondence-based methods. 
The former mainly consists of PointnetLK \cite{aoki2019pointnetlk} and its variants \cite{huang2020feature,xu2021omnet,li2021pointnetlk}. They usually extract two global features of given point clouds and directly estimate the transformation without correspondences. However, they face challenge on real-world low-overlap point clouds. 
The latter usually follows a pipeline of feature extraction, correspondence generation, and robust model fitting. 
Recently, benefiting from great progress in deep learning, many learning-based methods were proposed. 
FCGF \cite{ref8} first utilized sparse convolution \cite{ref36} to extract deep descriptors for point cloud registration, outperforming a series of hand-craft descriptors at that time. 
D3Feat \cite{ref9} first provided a network for extracting keypoints and descriptors simultaneously. 
Predator [5] introduced an overlap attention to estimate overlap region, improving registration effectively. 
YOHO \cite{ref6} designed an ensemble model to achieve SO(3)-equivariant and proposed modified RANSAC with lower complexity. 
Inspired by recent success in transformer \cite{ref16,ref17,ref18}, many transformer-based methods \cite{ref19,ref21} were proposed, taking long-dependency into consideration. 
However, most works inevitably tend to generate wrong correspondences facing repeatable and ambiguous structures due to reliance on geometric information. 
In this work, we incorporate extra information learned from multi-modality data to mitigate it. 

\subsection{Knowledge transfer between point cloud and image}
Fusion of point cloud and image \cite{ref10,ref22,ref23}, which contains complementary information, typically improves performance. 
FFB6D \cite{ref22} designed a bidirectional network to effectively fuse features extracted from different modalities and achieved great performances in 6D pose estimation. 
IMFNet \cite{ref10} successfully boosted registration by utilizing a cross-attention to fuse geometric information from point cloud and semantic information from image. 
However, all above methods need spatial and temporal calibration and synchronization for sensors, which limits their practical applications. 
To reserve the improvement while obtaining more flexible inference, many works attempted to transfer knowledge learned from multiple modalities to a single modality. 
They are roughly divided into two technical routes: network pre-training and knowledge distillation. 
The former commonly incorporates a network pre-trained on different modalities and utilizes knowledge in pre-trained weights during inference. 
For example, many monocular 3D detection methods \cite{ref24,ref25} leveraged depth estimation to obtain a pre-trained network and utilized it to produce pseudo-lidar to improve monocular detection. 
The latter achieves improvement by setting a network trained by multi-modality data as a teacher \cite{ref26,ref27}. 
For instance, S2M2-SSD \cite{ref26} trained a single-modality network to learn from a multi-modality network to obtain comparable performance but took only single-modality input at inference. 
In this paper, we choose the former route to boost registration. We also tried using knowledge distillation but failed, more details about our attempt are illustrated in experiments.

\begin{figure*}[htp]
  \centering
  \includegraphics[width=1.0\textwidth]{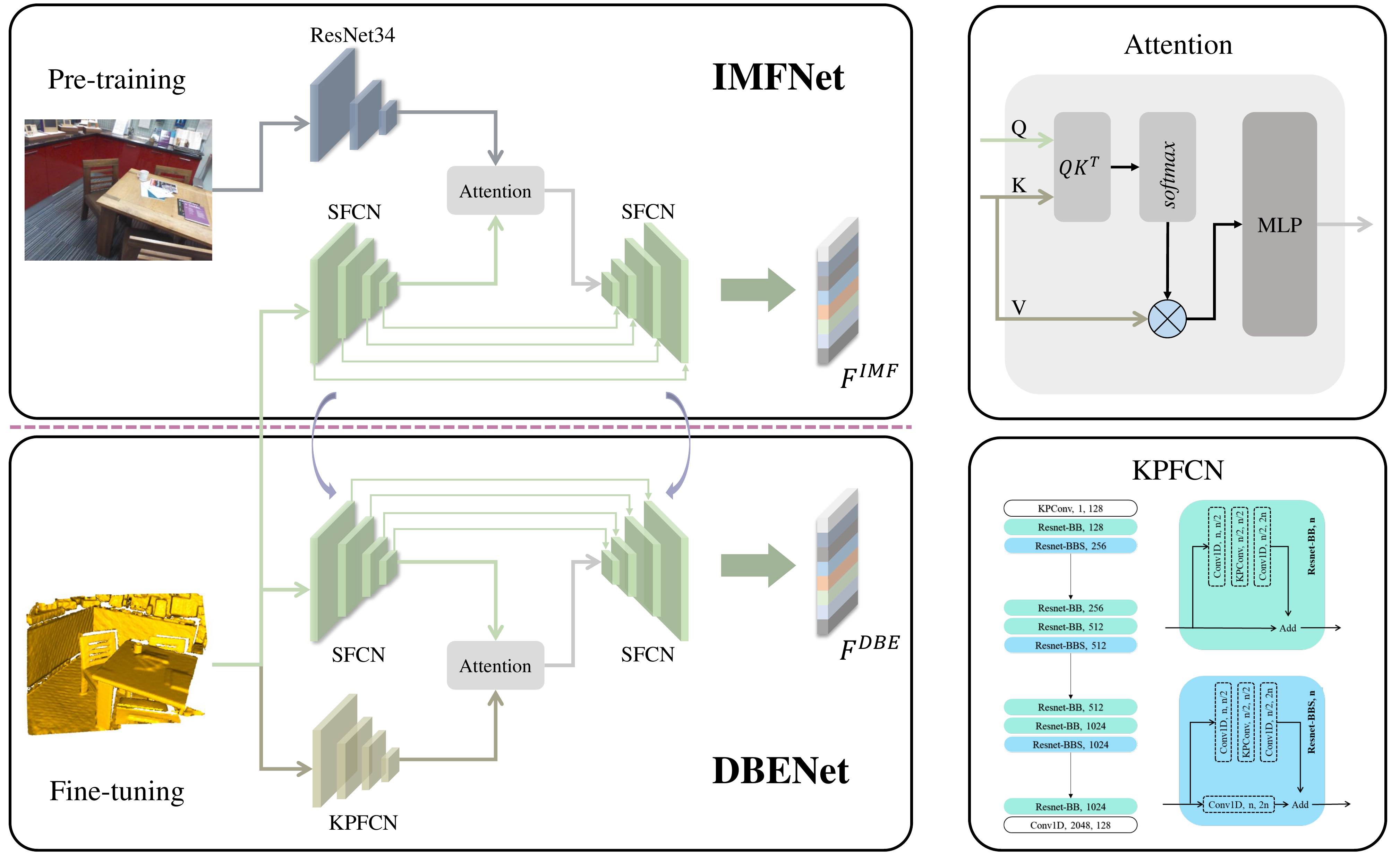}
  \caption{
  \textbf{The overall architecture of our proposed framework.} 
Our framework contains two dual-branch networks, IMFNet and DBENet. 
  IMFNet is an exiting model pre-trained on multiple modalities. 
  DBENet integrates two heterogeneous branches but only takes point clouds as input. 
  During fine-tuning, we begin with transferring the weights of SFCN in IMFNet to SFCN in DBENet. 
  Through such weights transfer, the knowledge learned from multiple modalities can be transferred to DBENet, and then we freeze the SFCN in DBENet and fine-tune DBENet using pure point clouds. 
  Benefiting from knowledge transfer and ensemble design for DBENet, our fine-tuned DBENet can achieve comparable or even better performance to IMFNet but with only single modality input at inference. 
  $F^{IMF}$ and $F^{DBE}$ both represent the feature matrix for point cloud registration. 
  }
  \label{fig2}
\end{figure*}

\section{Method}
Figure \ref{fig2} shows the pipeline of our framework. 
Our framework consists of two models, IMFNet \cite{ref10} and DBENet. 
IMFNet is an existing model pre-trained on multiple modalities, which contains knowledge we want to transfer. 
DBENet is our proposed \textbf{D}ual-\textbf{B}ranch \textbf{E}nsemble network, which receives the transferred knowledge. 
During fine-tuning, we begin with transferring the pre-trained weights of SFCN in IMFNet to the SFCN in DBENet. 
Then we utilize pure point clouds to fine-tune the KPFCN and attention module in DBENet.
At inference, our DBENet only takes point clouds as input and can achieve high performance benefiting from both transferred knowledge and ensemble design. 
The details of each component are introduced in the following subsections.

\subsection{IMFNet}
Since IMFNet \cite{ref10} has a dual-branch architecture, which separates the networks for two different modalities, we choose it as our pre-trained network. 
The architecture of IMFNet is shown in Figure \ref{fig2}, which consist of 3 components: encoder, cross-attention and decoder. 

The encoder contains two branches, which encodes features from point cloud and image, respectively. 
The branch for point cloud i.e. sparse fully convolution network (SFCN) is implemented by sparse convolution \cite{ref36} and has the same architecture as the encoder in FCGF \cite{ref8}. 

For an input point cloud $P \in R^{N \times 3}$, it gradually abstracts it and extracts high-level features $F_P \in R^{N^{\prime} \times 256}$, where $N^{\prime}$ denotes the number of abstracted points. 
The branch for image is a ResNet34 \cite{ref28}. 
It takes the corresponding image $I \in R^{H \times W \times 3}$ as input and outputs the flatten features $F_I \in R^{M^{\prime} \times 128}$, where $M^{\prime} = H/8*W/8$ denotes the number of subsampled pixels. 
After obtaining features from two different modalities, IMFNet use a scale dot-product attention \cite{ref16} as cross-attention module to fuse them to more reliable features $F^{IMF}_{fused} \in R^{N^{\prime} \times 256}$:
\begin{equation}
F^{IMF}_{{fused }}=F_P+M L P\left(\operatorname{softmax}\left(\frac{Q K^T}{\sqrt{d}}\right) V\right)
\end{equation}
where $Q=W_QF_P \in R^{N^{\prime} \times d}$ denotes the query matrix, while $K=W_KF_I \in R^{M^{\prime} \times d}$ and $V=W_VF_I \in R^{M^{\prime} \times d}$ denote the key matrix and the value matrix, $W_*$ denotes the learnable linear transformation mapping $F_*$ to $d$ dimension.  
Afterward, the fused features are gradually upsampled in U-net \cite{ref30} manner to output point-wise features $F^{IMF}$ for registration. 

\subsection{DBENet}
Our DBENet also contains two different branches, but two branches take the same point clouds as input. 
DBENet is more powerful than a single SFCN because it intergrates a new branch based on kernel point convolution. 
As mentioned in \cite{ref31}, voxel-based methods e.g. sparse convolution has advantage of extracting coarse-grain features, while point-based methods such as PointNet++ \cite{ref37}, KPConv \cite{ref29} has advantage of capturing fine-grain features. 
Therefore, we preserve the dual-branch architecture in IMFNet and replace the branch for image with a KPFCN encoder for point cloud to further improve the features extracted by sparse convolution network.

The details of KPFCN in DBENet are also shown in Figure \ref{fig2}. 
It consists of many KPConv-based blocks. 
Given a input point cloud $P \in R^{N \times 3}$ with features $F_{\text {in }} \in R^{N \times 1}=[1,1 \ldots 1]^T$, KPFCN gradually encodes spatial information into features and downsamples the point cloud. 
The final outputs of KPFCN are high-level features $F_P^{k p} \in R^{N^{\prime} \times 128}$, which have the similar dimension with flatten image features $F_I \in R^{M^{\prime} \times 128}$. 
We also use the same cross-attention module to fuse the features $F_P^{k p} \in R^{N^{\prime} \times 128}$ extracted by KPFCN with the features $F_P \in R^{N^{\prime} \times 256}$ extracted by the SFCN. 
After that, a decoder gradually upsamples the fuesd features $F^{DBE}_{{fused }} \in R^{N^{\prime} \times 256}$ and normalizes the features $F^{DBE} \in R^{N \times 32}$ in the last layer as outputs for registration. 
We can use $F^{DBE} \in R^{N \times 32}$ to generate putative correspondences and then use a robust estimator such as RANSAC \cite{ref7} to achieve registration during inference.

\subsection{Fine-tuning}
As mentioned above, point cloud contains rich geometric information and image texture contains rich semantic information, making feature extractor pre-trained on multiple modalities more distinctive. 
Therefore, we first transfer the knowledge contained in pre-trained model to our DBENet. 

Since our DBENet has a same SFCN with IMFNet and SFCN in IMFNet contains knowledge learned from multiple modalities, we can achieve knowledge transfer by directly transferring the network weights. 
Specifically, we replace the randomly initialized weights in DBENet with the pre-trained weights of SFCN in IMFNet and then freeze the transferred weights and fine-tune the weights of KPFCN and cross-attention module. 


We utilize the hardest contrastive loss \cite{ref8} to fine-tune our DBENet. 
Given a point cloud pair, we first sample some points to generate putative correspondences by feature matching and build the positive pair set $\mathcal{P}$ and the negative pair set $\mathcal{N}$. 
The positive pair set $\mathcal{P}$ contains feature pairs of the putative correspondences with limited residual error under the ground truth transformation, while the negative pair set $\mathcal{N}$ mines the hardest negatives $\hat{F}^{DBE}_i,\hat{F}^{DBE}_j$ for $(F^{DBE}_i,F^{DBE}_j)$ in $\mathcal{P}$ and remove false negatives. 
The loss function is formulated as: 
\begin{equation}
\begin{aligned}
\mathcal{L} &=\sum_{(i, j) \in \mathcal{P}}\{\left[D\left(F^{DBE}_i, F^{DBE}_j\right)-m_p\right]_{+}^2 /|\mathcal{P}|\\
&+\lambda_n I_i\left[m_n-\min _{k \in \mathcal{N}} D\left(F^{DBE}_i, F^{DBE}_k\right)\right]_{+}^2 /\left|\mathcal{P}_i\right| \\
&+\lambda_n I_j\left[m_n-\min _{k \in \mathcal{N}} D\left(F^{DBE}_j, F^{DBE}_k\right)\right]_{+}^2 /\left|\mathcal{P}_j\right|\}
\end{aligned}
\end{equation}
where $D(\cdot)$ denotes the Euclidean distance, $I[\cdot]$ is an indicator function that return 1 if the condition is satisfied and 0 otherwise, $m_p$ and $m_n$ denote the margins for positive and negative pairs, $\left|\mathcal{P}_i\right|$ denotes the number of valid hardest negatives for the first items in $\mathcal{P}$ and $\left|\mathcal{P}_j\right|$ for the second items, $\lambda_n$ denotes the weight to balance positive loss and negative loss, we set $m_p =0.1, m_n =1.4, \lambda_n = 0.5$ in fine-tuning.

\section{Experiment}
We evaluate our method by comparing it with many state-of-the-art methods on widely-used public datasets including 3DMatch \cite{ref11}, 3DLoMatch \cite{ref5}, and KITTI \cite{ref12}. 
The following sections are organized as follows. 
First, we illustrate our experimental settings including implementation details and evaluation metrics in section \ref{sec41}. 
Next, we conduct experiments on indoor datasets, 3DMatch and 3DLoMatch in section \ref{sec42} and \ref{sec43}. 
We also implement an experiment on outdoor dataset KITTI in section \ref{sec44}. To further understand our method, we conduct comprehensive ablation studies in section \ref{sec45}.

\subsection{Experimental settings}\label{sec41}
\textbf{Implementation: }
We implement our networks in Pytorch \cite{ref32}. 
For IMFNet \cite{ref10}, we directly use the pre-trained weights provided in \url{https://github.com/XiaoshuiHuang/IMFNet}. 
We use the ADAM optimizer \cite{kingma2014adam} with an initial learning rate of 0.1 to fine-tune our DBENet for 10 epochs and the batch size for fine-tuning is set to 2. 
All the experiments are conducted on a single GTX 1080ti graphic card with Intel Core i7-7800X CPU.

\textbf{Evaluation metrics:}
To quantitatively compare our method with other state-of-the-art methods, we select several widely-used evaluation metrics to evaluate performance. 

For 3DMatch \cite{ref11} and 3DLoMatch \cite{ref5} benchmark, the most widely-used evaluation metrics are: Feature-Match Recall (FMR), Inlier-Ratio (IR), and Registration Recall (RR). 
FMR measures the percentage of pairs that have $>5\%$ inlier correspondences with 10cm residual under the ground truth transformations. 
IR describes the percentage of inlier correspondences among all the putative correspondences generated by feature matching. 
And RR directly shows the percentage of successfully registered pairs. In these two benchmarks, we consider the registration with $E_{RMSE}<0.2m$ as a successfully registered pair. 
$E_{RMSE}$ denotes the error metric between an unaligned point cloud pair $\{i,j\}$:
\begin{equation}
E_{R M S E}=\sqrt{\frac{1}{\left|\Omega^*\right|} \sum_{\left(x^*, y^*\right) \in \Omega^*}\left\|\hat{T}_{i, j} x^*-y^*\right\|^2}
\end{equation}
where $\hat{T}_{i, j}$ represents the estimated transformations for point cloud pair $\{i,j\}$, $\Omega^*$ represents the set containing all the inlier correspondences, $x^*$ and $y^*$ represent the 3D coordinates in it. 

For KITTI [12] benchmark, the most widely-used evaluation metrics are: Relative Translation Error (RTE), Relative Rotation Error (RRE) and Success rate (Success). 
RTE is defined as RTE$=\left|\hat{t}-t^*\right|$, where $\hat{t}$ denotes the estimated translation and $t^*$ denotes the ground truth translation. 
RRE is defined as RRE$=\arccos \left(\left(\operatorname{Tr}\left(\hat{R}^T R^*\right)-1\right) / 2\right)$, where $\hat{R}$ denotes the estimated rotation and $R^*$ denotes the ground truth rotation. 
Success rate measures the percentage of registered pairs with RTE$<2m$ and RRE$<5^{\circ}$.

\subsection{Experiment on 3DMatch}\label{sec42}
3DMatch \cite{ref11} is a well-known dataset, which consists of 62 scenes. 
Here we follow the official split \cite{ref5} to divide the dataset into 46 scenes for training, 8 scenes for validation and 8 scenes for test. 
We compare our methods with many state-of-the-art methods. 
Besides IMFNet \cite{ref10}, all the competitors are trained and tested on point clouds and without refinement during test, while IMFNet is trained and tested using two different modalities. 

The results of our method and competitors are shown in Table \ref{table1}. 
Our method surprisingly achieves new state-of-the-art on all evaluation metrics, and even outperforms the method using multiple modalities. 
Moreover, our method successfully boosts the performance of baseline i.e. FCGF \cite{ref8} by a large margin. More analysis on improvement is illustrated in ablation studies.

\begin{table}[]
\centering
\caption{Results on 3DMatch dataset.}
\setlength{\tabcolsep}{0.38cm}{
\begin{tabular}{l|ccc}
\hline
                        & FMR (\%)      & IR (\%)       & RR (\%)       \\ \hline
3DSN \cite{ref33}           & 94.7          & 36.0          & 78.4          \\
FCGF \cite{ref8}           & 95.2          & 56.8          & 85.1          \\
D3Feat \cite{ref9}          & 95.8          & 39.0          & 81.6          \\
Predator \cite{ref5}        & 96.6          & 61.0          & 88.3          \\
SpinNet \cite{ref34}       & 97.6          & 47.5          & 88.6          \\
YOHO \cite{ref6}           & 98.2          & 64.4          & 90.8          \\
CoFiNet \cite{ref35}       & 98.1          & 49.8          & 89.3          \\
GeoTransformer \cite{ref21} & 97.9          & 71.9          & 92.0          \\
REGTR \cite{ref20}       & -             & -             & 92.0          \\
Lepard \cite{ref19}     & 98.3          & 55.5          & 93.5          \\
IMFNet \cite{ref10}        & \textbf{98.6} & 85.5          & 93.4          \\
Ours                    & \textbf{98.6} & \textbf{86.1} & \textbf{93.8} \\ \hline
\end{tabular}\label{table1}}
\end{table}

\subsection{Experiment on 3DLoMatch}\label{sec43}
3DLoMatch \cite{ref5} is a much more challenging benchmark, which consists of low-overlap ratio (10\%-30\%) point cloud pairs from 3DMatch. 
All the competitors tested on 3DLoMatch are previously trained on 3DMatch. 

The results on 3DLoMatch are shown in Table \ref{table2}. 
It’s observed that our method only achieves state-of-the-art on IR metric. 
This is because the state-of-the-art methods such as GeoTransformer \cite{ref21} are trained using overlap-aware loss, which provides extra supervision for network to learn to estimate overlap regions. 
Although these overlap-based methods have a natural advantage when facing low-overlap point cloud pairs, our method still shows a competitive performance even outperforms Predator \cite{ref5}, which contains an overlap attention. 
Among methods without overlap supervision such as YOHO \cite{ref6} and SpinNet \cite{ref34}, our method is a cost-effective choice due to its simpler implementation and comparable performance. 
Moreover, our method achieves comparable performance with IMFNet \cite{ref10}, but our method only takes point clouds as input during inference, avoiding the challenge of calibration and synchronization for sensors. 
As on 3DMatch benchmark, our method also boosts baseline i.e. FCGF by a large margin, indicating the effectiveness of our method. 

\begin{table}[]
\centering
\caption{Results on 3DLoMatch dataset.}
\setlength{\tabcolsep}{0.38cm}{
\begin{tabular}{l|ccc}
\hline
                        & FMR (\%)      & IR (\%)       & RR (\%)       \\ \hline
3DSN \cite{ref33}           & 63.6          & 11.4          & 33.0          \\
FCGF \cite{ref8}         & 76.6          & 21.4          & 40.1          \\
D3Feat \cite{ref9}         & 67.3          & 15.0          & 46.9          \\
Predator \cite{ref5}      & 78.6          & 38.0          & 56.7          \\
SpinNet \cite{ref34}     & 75.3          & 20.5          & 59.8          \\
YOHO \cite{ref6}       & 79.4          & 25.9          & 65.2          \\
CoFiNet \cite{ref35}     & 83.1          & 24.4          & 67.5          \\
GeoTransformer \cite{ref21} & \textbf{88.3} & 43.5          & \textbf{75.0} \\
REGTR \cite{ref21}       & -             & -             & 64.8          \\
Lepard \cite{ref19}     & 84.5          & 26.0          & 69.0          \\
IMFNet \cite{ref10}       & 80.3 & 46.6          & 65.9          \\
Ours                    & 80.3 & \textbf{47.7} & 65.0 \\ \hline
\end{tabular}\label{table2}}
\end{table}

\subsection{Experiment on KITTI}\label{sec44}
KITTI \cite{ref12} is one of the most well-known datasets for autonomous driving, which contains 3D point clouds captured by LiDAR. 
We follow previous settings in \cite{ref10} and divide 11 sequences (0-10) of the odometry dataset into training set (0-5), validation set (6-7) and test set (8-10). 
Referring to the experimental settings in \cite{ref10}, we report both the averages and standard deviations for RTE and RRE. 

The results are shown in Table \ref{table3}. 
Although our method does not achieve the best performance, it still achieves comparable performance to the method trained on multiple modalities. 
Moreover, our method also boosts the performance of FCGF \cite{ref8}, which serves as a baseline in our experiments. This also reflects the effectiveness of our method.

\begin{table}[]
\centering
\caption{Results on KITTI dataset.}
\setlength{\tabcolsep}{0.12cm}{
\begin{tabular}{l|ccccc}
\hline
                 & RTE (cm) & STD (cm) & RRE (°) & STD (°)       & Success (\%)   \\ \hline
FCGF \cite{ref8}   & 6.47     & 6.07     & \textbf{0.23}    & 0.23          & 98.92          \\
D3Feat \cite{ref9}  & 6.90     & 0.30     & 0.24    & 0.06          & \textbf{99.81} \\
Predator \cite{ref5} & 6.80     & -        & 0.27    & -             & 99.80          \\
SpinNet \cite{ref34} & 9.88     & 0.50     & 0.47    & 0.09          & 99.10          \\
IMFNet \cite{ref10} & \textbf{5.77}     & \textbf{0.27}     & 0.37    & \textbf{0.01} & 99.28          \\
Ours             &     5.98  &  0.29  &   0.42      & \textbf{0.01} & 99.10          \\ \hline
\end{tabular}\label{table3}
}
\end{table}

\subsection{Ablation studies}\label{sec45}
To further understand our work, we conduct several ablation studies. 
First, we conduct an experiment to illustrate the effectiveness of multi-modality pre-training and our ensemble design. 
Second, we empirically demonstrate our decision to use the aforementioned fine-tuning strategy. 
Finally, we briefly discuss another knowledge transfer method, knowledge distillation and illustrate our attempt. 
All the ablation studies are conducted on 3DMatch dataset. 

\textbf{Effectiveness of multi-modality pre-training and model ensemble.} 
As mentioned in previous paragraph, multi-modality data is more informative, which helps network to learn a more distinctive descriptor. 
We verify it by replacing the weights in an original FCGF \cite{ref8} with that in IMFNet \cite{ref10}. 
The comparison is shown in line 2 and 3 in Table \ref{table4}. 
Here we use P to denote using the weights trained on pure point cloud, and use P+I to denote using the weights coming from pre-trained IMFNet.  
It’s observed that the performance of FCGF using multi-modality pre-trained weights performs better, which verifies our proposal. 

As illustrated in \cite{ref31}, voxel-based feature and point-based feature are commonly complementary. 
The former has advantages in learning structure, while the later has advantages in capturing details. 
Therefore, the ensemble model which has heterogeneous branches tends to perform better. 
Line 2 and 4 in Table \ref{table4} also verify it. 
We compare the baseline i.e. FCGF with a DBENet trained without using transferred weight. 
The DBENet trained from scratch also improves the performance of the baseline, indicating the effectiveness of our ensemble design.  
All in all, both the multi-modality pre-training and model ensemble help our method achieve outstanding performance on point cloud registration.

\begin{table}[]
\centering
\caption{Ablation on multi-modality pre-training and model ensemble. }
\setlength{\tabcolsep}{0.5cm}{
\begin{tabular}{l|ccc}
\hline
               & FMR (\%)      & IR (\%)       & RR (\%)       \\ \hline
FCGF (P)       & 95.2          & 56.8          & 85.1          \\
FCGF (P+I)     & 97.7          & 85.4          & 92.9          \\
Ours (scratch) & 96.7          & 83.3          & 92.1          \\
Ours           & \textbf{98.6} & \textbf{86.1} & \textbf{93.8} \\ \hline
\end{tabular}\label{table4}}
\end{table}

\textbf{Ablation on fine-tuning strategies.} Pre-training and then fine-tuning is a widely-used paradigm. 
Generally speaking, freezing encoder and fine-tuning other modules is the most conventional choice. 
We also attempt to freeze other pre-trained modules and find that freezing encoder and decoder and then fine-tuning attention and KPFCN is the best choice as shown in Table \ref{table5}.

\begin{table}[]
\centering
\caption{Ablation on fine-tuning strategies. 
$\checkmark$  denotes freezing the pre-trained weight.}
\begin{tabular}{ccc|ccc}
\hline
Encoder & Attention & Decoder & FMR (\%) & IR (\%) & RR (\%) \\ \hline
$\checkmark$       &           &         & 98.5    &  85.6  & 93.1    \\
$\checkmark$       & $\checkmark$     &         & 98.3 &     85.8   & 93.1    \\
$\checkmark$       &$\checkmark$       & $\checkmark$     & 98.3 &   \textbf{86.2}     & 93.3    \\
$\checkmark$        &      & $\checkmark$      & \textbf{98.6}  &  86.1     & \textbf{93.8}    \\ \hline
\end{tabular}\label{table5}
\end{table}

\textbf{Discussion on knowledge transfer.} 
As mentioned in related works, there exists the other methods to transfer knowledge learned from multiple modalities to a single modality. 
Therefore, we designed a knowledge distillation strategy to achieve it. 

Since point cloud is not as regular as image, the number of input points is uncertain and the extracted high-level features are unordered, making the features extracted by KPFCN impossible to approximate the features extracted by ResNet34 by a direct supervision. 
Therefore, we designed a point-to-point teacher-student loss in the output of the cross-attention module to avoid pixel-to-point direct distillation.

We hope the teacher i.e. pre-trained IMFNet to guides the student i.e. DBENet, so that the fused features $F_{{fused }}^{DBE}$ extracted by student can simulate the fused features $F_{{fused }}^{IMF}$ extracted by teacher. 
To make fused features as consistent as possible, we add a point-to-point KL loss during fine-tuning:
\begin{equation}
\mathcal{L}_{K D}=\frac{1}{256 * N^{\prime}} \sum_{i=1}^N \sum_{j=1}^{256} \phi\left(F_{{fused }, i, j}^{I M F}\right) \cdot \log \left[\frac{\phi\left(F_{{fused }, i, j}^{I M F}\right)}{\phi\left(F_{{fused }, i, j}^{D B E}\right)}\right]
\end{equation}
where $\phi(\cdot)$ denotes a channel-wise softmax function \cite{hinton2015distilling} with temperature $T=1$. 

\begin{table}[ht]
\centering
\caption{Ablation on knowledge transfer strategies. }
\setlength{\tabcolsep}{0.4cm}{
\begin{tabular}{cc|ccc}
\hline
Pre-training & KD & FMR (\%)      & IR (\%)       & RR (\%)       \\ \hline
             &    & 96.7          & 83.3          & 92.1          \\
             & $\checkmark$  & 93.1          & 78.8          & 88.2          \\
$\checkmark$           & $\checkmark$ & 98.3          & 85.8          & 93.3          \\
$\checkmark$            &    & \textbf{98.6} & \textbf{86.1} & \textbf{93.8} \\ \hline
\end{tabular}\label{table6}}
\end{table}
\vspace{0.5cm}

\begin{figure}[htp]
  \centering
  \includegraphics[width=0.5\textwidth]{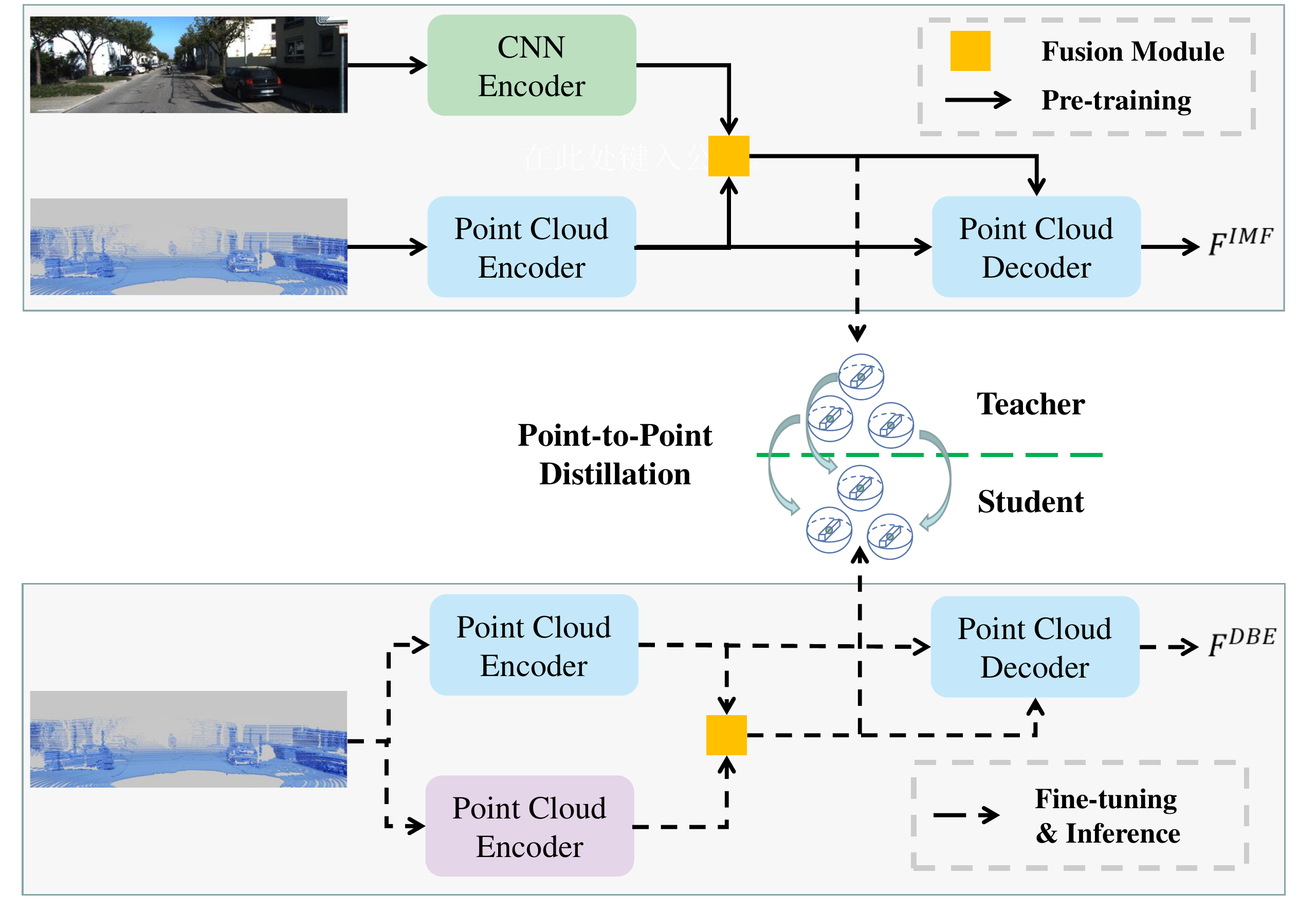}
  \caption{
  Multi-modality knowledge distillation for point cloud registration. 
  }
  \label{fig3}
\end{figure}

The results are shown in Table \ref{table6}. 
Pre-training denotes using pre-trained weights, KD denotes using an extra point-to-point teacher-student loss during fine-tuning. 
Although it sounds reasonable to add a loss to make the student features simulate the teacher's as much as possible to improve student, it fails in practical experiment and even has a negative impact. 
We infer that the large gap between two completely different modalities and network architectures makes it fail to distillate knowledge explicitly. 

We also attempted replacing KL loss with L1 loss, but get similar results. 
It can be seen that for heterogeneous cross-modality networks, pre-training may be a better way of knowledge transfer than distillation. 
Unified networks between different modalities such as transformer may \cite{ref16,radford2021learning} have the potential for knowledge transfer in terms of knowledge distillation because it may be able to narrow the gap caused by the difference in network architectures.

\section{Conclusion}
In this paper, we propose a method to transfer the multi-modality knowledge to boost the performance of point cloud registration. Our proposed method ensembles the pre-trained sparse convolution branch and point convolution branch, which can leverage the multi-modality knowledge and utilize only the point cloud modality during inference. The proposed method does not require the strict calibration and synchronization of multiple modalities during the inference. The experiments show that the ensemble model with multi-modality knowledge can significantly improve the registration accuracy and even outperform the multi-modality model.

\section*{ACKNOWLEDGMENT}
This work was supported by the National Natural Science Foundation of China under Grant 62076070. 


\newpage
\bibliographystyle{IEEEtrans}
\bibliography{ref}

\end{document}